%% file: vcnn_iccv15.tex
\documentclass[10pt,twocolumn,letterpaper]{article}
\usepackage{iccv}
\usepackage{times}
\usepackage{epsfig}
\usepackage{wrapfig}
\usepackage{graphicx}
\usepackage{amsmath}
\usepackage{amssymb}
\usepackage{array}
\graphicspath{{./images/}}

\usepackage[pagebackref=true,breaklinks=true,letterpaper=true,colorlinks,bookmarks=false]{hyperref}

 \iccvfinalcopy 

\def\iccvPaperID{2163} 

\newcommand{\rebuttal}[1]{\textcolor{black}{#1}}

\ificcvfinal\fi
\begin{document}


\title{Learning Deep Object Detectors from 3D Models}

\author{Xingchao Peng,   Baochen Sun,   Karim Ali,   Kate Saenko \\  
University of Massachusetts Lowell\\
{\tt\small \{xpeng,bsun,karim,saenko\}@cs.uml.edu}
}

\maketitle

\begin{abstract}
Crowdsourced 3D CAD models are becoming easily accessible online, and can
potentially generate an infinite number of training images for almost any object category. We show that augmenting the training data of contemporary Deep Convolutional Neural Net (DCNN) models with such synthetic data can be effective, especially when real training data is limited or not well matched to the target domain.
Most freely available CAD models capture 3D shape but are often missing other low level cues, such as realistic object texture, pose, or background. In a detailed analysis, we use synthetic CAD-rendered images to probe the ability of DCNN to learn without these cues, with surprising findings. In particular, we show that when the DCNN is fine-tuned on the target detection task, it exhibits a large degree of invariance to missing low-level cues, but, when pretrained on generic ImageNet classification, it learns better when the low-level cues are simulated.
We show that our synthetic DCNN training approach significantly outperforms
previous methods on the PASCAL VOC2007 dataset when learning in the
few-shot scenario and improves performance in a domain shift scenario on the Office benchmark.

\end{abstract}

\input{introduction}

\input{related}

\input{method}

\input{experiments}

\input{conclusion}

{\small
\bibliographystyle{ieee}
\bibliography{iclr2015}
}

\end{document}

%% file: introduction.tex
\section{Introduction}
\label{introduction}

Deep CNN models achieve state-of-the-art performance on
object detection, but are heavily dependent on large-scale training
data. Unfortunately, labeling images for detection is extremely time-consuming,
as every instance of every object must be marked with a bounding box. Even the
largest challenge datasets provide a limited number of annotated categories,
e.g., 20 categories in PASCAL VOC~\cite{PASCAL}), 80 in COCO~\cite{mscoco}, and
200 in ImageNet~\cite{ImageNet}. But what if we wanted to train a detector for a
novel category? It may not be feasible to compile and annotate an extensive
training set covering all possible intra-category variations.

\begin{figure} 
\centering
\includegraphics[width=0.8\linewidth]{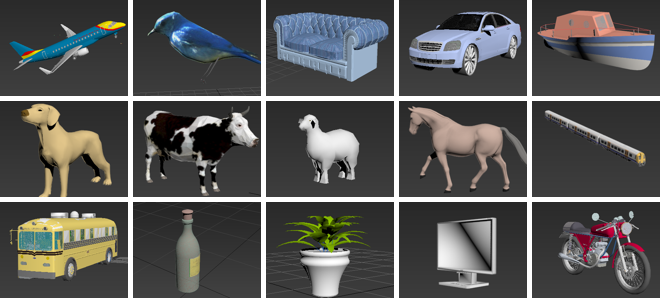}
\caption{\small We propose to train few-shot object detectors for real images by augmenting the training data with synthetic images generated from freely available non-photorealistic 3D CAD models of objects collected from \url{3dwarehouse.sketchup.com}.}
\label{fig:cad}
\vspace{-0.2in}
\end{figure}

\begin{figure*} \centering
\includegraphics[width=0.9\linewidth]{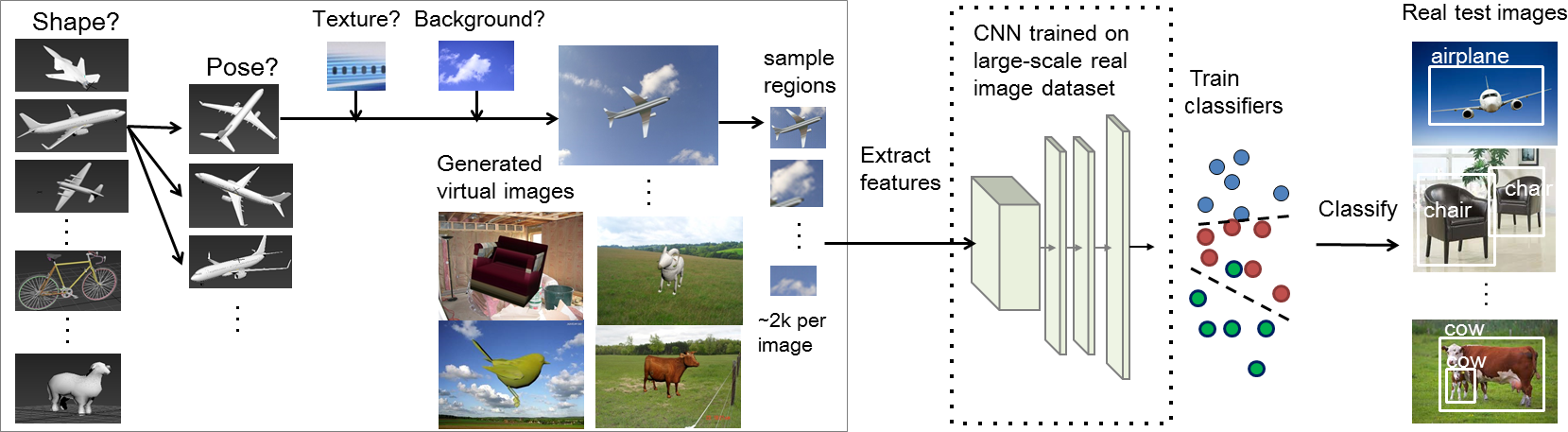}
\caption{\small Can we learn deep detectors for real images from non-photorealistic 3D CAD models? We explore the invariance of deep features to missing low-level cues such as shape, pose, texture and context, and propose an improved method for learning from synthetic CAD data that simulates these cues.}
\vspace{-0.2in}
\label{fig:overview}
\end{figure*}

We propose to bypass the expensive collection and annotation of real images by
using freely available 3D CAD models to automatically generate synthetic 2D
training images (see Figure~\ref{fig:overview}). Synthetic data augmentation has been used successfully in the
past to add 2D affine transformations to training images~\cite{alexnet},
recognize text~\cite{syntext}, and even train detectors for a handful of
categories such as cars~\cite{stark_bmvc10}. However it has not yet been
demonstrated for detection of many categories  with modern DCNNs. \cite{BMVC}
trained object detectors for 31 categories on synthetic CAD images, but used a histogram-of-oriented (HOG) gradient model (DPM~\cite{dpm}), which is significantly less powerful than DCNNs on object classification~\cite{alexnet} and detection~\cite{Overfeat,RCNN,vgg}.

The main challenge in training with freely available CAD models is that they capture the 3D shape of the object, but frequently lack other low-level cues, such as object texture, background, realistic pose, lighting, etc. 
\cite{BMVC} used a simple rendering of objects with uniform gray texture and a white background, and showed that HOG-based models learn well from such data, as they are invariant to color and texture and mostly retain the overall shape of the object.
However, DCNN visualizations have shown that they retain color, texture and mid-level patterns. It is
therefore unknown if they would tolerate the lack of such low-level cues in training images, or if a more sophisticated rendering process that simulates these cues is needed.

To investigate how missing low-level cues affect DCNNs' ability to learn object detectors, we study the precise nature of their ``cue invariances''. For a given object category, a DCNN maps the low-level cues contained in the image (shape, texture) to high-level category information (cat, car) represented by top layer activations (e.g. fc7 in AlexNet~\cite{alexnet}).
We define ``cue invariance'' to be the ability of the network to extract the equivalent high-level category information despite missing low-level cues. We expect the network to learn different invariances depending on the task it was trained on.

Quantifying such invariances could help better understand DCNN models and impove transfer to
new domains, e.g., to non-photorealistic data. A small number of papers have
started looking at this problem~\cite{lenc14arXiv,yosinskiNIPS14,mahendran14arXiv}, but many open questions remain, such as: are DCNNs invariant to object color? Texture? Context? 3D pose? Is the invariance transferable to new tasks?  


With the help of images synthetically rendered from 3D models, we design a series of experiments to ``peer into the depths'' of DCNNs and analyse their invariance to cues, including ones that are difficult to isolate using real 2D image data. We make surprising discoveries regarding the representational power of deep features. In particular, we show that they encode far more complex invariances to cues such as 3D pose, color, texture and context than previously accounted for. We also quantify the degree to which the learned invariances are specific to the training task.

Based on our analysis, we propose a method for zero- or few-shot learning of
novel object categories that generates synthetic 2D data using 3D models and a few texture and scene images related to the category. An
advantage of our approach is that it drastically reduces the amount of human supervision over traditional bounding-box labeling methods.
This could greatly expand available sources of
visual knowledge and allow learning 2D detectors from the millions of CAD models
available on the web. We present experiments on the PASCAL VOC 2007 detection
task and show that when training data is missing or limited for a novel
category, our method outperforms both training on real data and the synthetic 
method of~\cite{BMVC}. We also demonstrate the advantage of our approach in the
setting when the real training data comes from a different domain than target
data using the Office~\cite{office} benchmark.


To summarize, our contributions are three-fold:
\vspace{-0.5em}
\begin{itemize}
\setlength\itemsep{-0.5em}
\item we gain new and important insights into the cue invariance of DCNNs through the use of synthetic data,
\item we show that synthetic training of modern large-scale DCNNs improves
detection performance in the few-shot and dataset-bias scenarios,
\item we present the largest-scale evaluation of synthetic CAD training of object detectors to date.
\end{itemize}


%% file: related.tex
\section{Related Work} \label{related}

\vspace{-1em}
\noindent\paragraph{Object Detection.} ``Flat'' hand-designed representations (HOG,
SIFT, etc.) have dominated the object detection literature due to their
considerable invariance to factors such as illumination, contrast and small
translations. In combination with discriminative classifiers such as linear SVM,
exemplar-based~\cite{exemplarsvm} or latent SVM~\cite{dpm}, they had proved
powerful for learning to localize the global outline of an object. More
recently, convolutional neural networks~\cite{lecun89} have overtaken flat
features as clear front-runners in many image understanding tasks, including
object detection.  DCNNs learn layered features starting with familiar pooled
edges in the first layer, and progressing to more and more complex patterns with
increasing spatial support. Extensions to detection have included sliding-window
CNN~\cite{Overfeat} and Regions-CNN (RCNN)~\cite{RCNN}.

\vspace{-1em}
\noindent\paragraph{Understanding Deep CNNs.} There has been increasing interest in
understanding the information encoded by the highly nonlinear deep
layers. \cite{zeiler-arxiv-2013} reversed the computation to find image patches
that most highly activate an isolated neuron.  A detailed study of what happens
when one transfers network layers from one dataset to another was presented by
\cite{yosinskiNIPS14}.  \cite{mahendran14arXiv} reconstruct an image from one
layer's activations, using image priors to recover the natural statistics
removed by the network filters. Their visualizations confirm that a
progressively more invariant and abstract representation of the image is formed
by successive layers, but they do not analyse the nature of the
invariances. Invariance to simple 2D transformations (reflection, in-plane
rotation) was explored by \cite{lenc14arXiv}. In this paper, we study more
complex invariances by ``deconstructing'' the image into 3D shape, texture, and
other factors, and seeing which specific combinations result in high-layer representations
discriminant of object categories.

\vspace{-1em}
\noindent\paragraph{Use of Synthetic Data.} The use of synthetic data has a
longstanding history in computer vision. Among the earliest attempts,~\cite{nevatia77}
used 3D models as the primary source of information to build object models. More recently,
\cite{stark_bmvc10,liebelt_cvpr10,sun_cvpr09} used 3D CAD models as their only
source of labeled data, but limited their work to a few categories like cars and
motorcycles. \cite{pinto2011comparing} utilized synthetic data to probe
invariances for features like SIFT, SLF, etc.  In this paper, we generate
training data from crowdsourced 3D CAD models, which can be noisy and
low-quality, but are free and available for many categories. We evaluate our
approach on all 20 categories in the PASCAL VOC2007 dataset, which is much
larger and more realistic than previous benchmarks. 

Previous works designed
special features for matching synthetic 3D object models to real image data
(\cite{liebelt_cvpr08}), or used HOG features and linear SVMs (\cite{BMVC}). We
employ more powerful deep convolutional images features and demonstrate their
advantage by directy comparing to \cite{BMVC}. The authors of~\cite{xiang_wacv14} use CAD
models and show results of both 2D detection and pose estimation, but train
multi-view detectors on real images labeled with pose. We avoid expensive manual
bounding box and pose annotation, and show results with
minimum or no real image labels. Finally, several approaches had used synthetic
training data for tasks other than object detection. For example,
\cite{syntext} recently proposed a synthetic text generation engine to
perform text recognition in natural scenes while~\cite{kostas14cvpr} proposed a
technique to improve novel-view synthesis for images using the structural
information from 3D models.



%% file: method.tex
\section{Approach}

Our approach learns detectors for objects with no or few training
examples by augmenting the training data with synthetic 3D CAD images. 
An overview of the approach is shown in Figure~\ref{fig:overview}. Given a set
of 3D CAD models for each object, it generates a synthetic 2D image training
dataset by simulating various low-level cues (Section~\ref{sec:cues}).
It then extracts positive and negative patches for each object from the synthetic images (and an optional small number of real images). Each patch is fed into a deep neural network that
computes feature activations, which are used to train the final classifier, as in the deep detection method of RCNN~\cite{RCNN} (Section~\ref{sec:rcnn}).
We explore the cue invariance of networks trained in different ways, as described in Section~\ref{sec:adapt}.

\subsection{Synthetic Generation of Low-Level Cues}
\label{sec:cues}

Realistic object appearance depends on many low-level cues, including
object shape, pose, surface color, reflectance, location and spectral
distributions of illumination sources, properties of the background scene,
camera characteristics, and others. We choose a subset of factors
that can easily be modeled using computer graphics techniques, namely,
\textbf{object texture, color, 3D pose and 3D shape}, as well as
\textbf{background scene texture and color}.

When learning a detection model for a new category with limited labeled
real data, the choice of whether or not to simulate these cues in
the synthetic data depends on the invariance of the representation. For example,
if the representation is invariant to color, grayscale images can be
rendered. We study the invariance of the DCNN representation to these
parameters using synthetic data generated as follows.



\paragraph{3D Models and Viewpoints}
Crowdsourced CAD models of thousands of objects are becoming freely available
online. We start by downloading models from \textit{3D Warehouse} by searching
for the name of the desired object categories.  For each category, around $5-25$ models were obtained for our experiments, and we explore the effect of varying intra-class shape by restricting the number of models in our experiments. The original poses
of the CAD models can be arbitrary (e.g., upside-down chairs, or tilted
cars). We therefore adjust the CAD models's viewpoint manually to $3$ or $4$
``views'' (as shown in Figure \ref{fig:overview}) that best represent
intra-class pose variance for real objects. Next, for each manually specified
model view, we generate several small perturbations by adding a random
rotation. Finally, for each pose perturbation, we select the texture, color and
background image and render a virtual image to include in our virtual training
dataset.  Next, we describe the detailed process for each of these factors.


\vspace{-1em}
\paragraph{Object/Background Color and Texture}
We investigate various combinations of color  and texture
cues for both the object and the background image. Previous work
by~\cite{BMVC} has shown that when learning detectors from virtual data using
HOG features, rendering natural backgrounds and texture was not helpful, and
equally good results were obtained by white background with uniform gray object
texture. They explain this by the fact that a HOG-based classifier is focused on
learning the ``outlines'' of the object shape, and is invariant to color and
texture. We hypothesise that the case is different for DCNN representations,
where neurons have been shown to respond to detailed textures, colors and
mid-level patterns, and explore the invariance of DCNNs to such factors.

Specifially, we examine the invariance of the DCNN representation to two types
of object textures: realistic color textures and uniform grayscale textures
(i.e., no texture at all). In the case of background scenes, we examine
invariance to three types of scenes, namely real-image color scenes, real-image
grayscale scenes, and a plain white background. Examples of our texture and
background generation settings are shown in Table~\ref{tab:color}.

In order to simulate realistic object textures, we use a small number ($5$ to
$8$ per category) of real images containing real objects and extract the
textures therein by annotating a bounding box. These texture images are then
stretched to fit the CAD models. Likewise, in order to simulate realistic
background scenes, we gathered about $40$ (per category) real images of scenes
where each category is likely to appear (e.g blue sky images for aeroplane,
images of a lake or ocean for boat, etc.) When generating a virtual image, we
first randomly select a background image from the available background pool, and
project it onto the image plane. Then, we select a random texture image from the
texture pool and map it onto the CAD model before rendering the object.

\subsection{Deep Convolutional Neural Network Features}
\label{sec:rcnn}

To obtain a deep feature representation of the images, we use the eight-layer
``AlexNet'' architecture with over 60 million parameters~\cite{alexnet}. This
network had first achieved breakthrough results on the
ILSVRC-2012~\cite{ilsvrc2012} image classification, and remains the most studied
and widely used visual convnet.  The network is trained by fully supervised
back-propagation (as in~\cite{lecun89}) and takes raw RGB image pixels of a
fixed size of 224 $\times$ 224 and outputs object category labels.  Each layer
consists of a set of neurons, each with linear weights on the input followed by
a nonlinearity.  The first five layers of the network have local spatial support
and are convolutional, while the final three layers are fully-connected to each
neuron from the previous layer, and thus include inputs from the entire image.

This network, originally designed for classification, was applied and fine-tuned
to detection in RCNN~\cite{RCNN} with impressive gains on the popular object detection
benchmarks. To adapt AlexNet for detection, the RCNN applied the network to each
image sub-region proposed by the Selective Search method
(\cite{selective-search}), adding a background label, and applied non-maximal
suppression to the outputs. Fine-tuning all hidden layers resulted in
performance improvements. We refer the reader to~\cite{RCNN} for more details.

\subsection{Analysing Cue Invariance of DCNN Features}
\label{sec:adapt}

Recall that we define ``cue invariance'' to be the ability of the network to extract the same high-level category information from training images despite missing low-level cues such as object texture. To test for this invariance, we create two synthetic training sets, one with and one without a particular cue. We then extract deep features from both sets, train two object detectors, and compare their performance on real test data. 
Our hypothesis is that, if the representation is invariant to the cue, then similar high-level neurons will activate whether or not that cue is present in the input image, leading to similar category-level information at training and thus similar performance. On the other hand, if the features are not invariant, then the missing cue will result in missing category information and poorer performance.
In this work, we extract the last hidden layer (fc7 of AlexNet) as the feature representation, since it has learned the most class-specific cue invariance.

As an example, consider the ``cat'' object class. If the network is invariant to cat texture, then it will produce similar
activations on cats with and without texture, i.e. it will ``hallucinate'' the
right texture when given a texureless cat shape. Then the detector will learn
cats equally well from both sets of training data. If, on the other hand, the network is not invariant to cat texture, then the feature distributions will differ, and the classifier trained on textureless cat data will perform worse.

We expect that the network will learn different cue invariances depending
on the task and categories it is trained on. For example, it may choose to focus on just the texture cue when detecting leopards, and not their shape or context, as their texture is unique.
To evaluate the effect of task-specific pre-training, we compare three different variants of the network:
1) one pre-trained on the generic ImageNet~\cite{ImageNet} ILSVRC 1000-way classification task ({\small IMGNET}); 2) the same network additionally fine-tuned on the PASCAL 20-category detection task ({\small PASC-FT}); and
3) for the case when a category has no or few labels, we fine-tune the {\small IMGNET} network on synthetic CAD data ({\small VCNN}).


To obtain the {\small VCNN} network, we fine-tune the entire network on the synthetic data by backpropagating the gradients with a lower learning rate. This has the effect of adapting the hidden
layer parameters to the synthetic data. It also allows the network to learn new
information about object categories from the synthetic data, and thus gain new
object-class invariances. We show that this is essential for good performance in the few-shot scenario. Treating the network activations as fixed features is inferior as most of the learning capacity is in the hidden layers, not the final classifier. 
We investigate the degree to which the presence of different low-level cues 
affects how well the network can learn from the synthetic data.

%% file: experiments.tex
\section{Experiments}

\subsection{Cue Invariance Results}
\begin{figure}[t]
\centering
\includegraphics[width=0.85\linewidth]{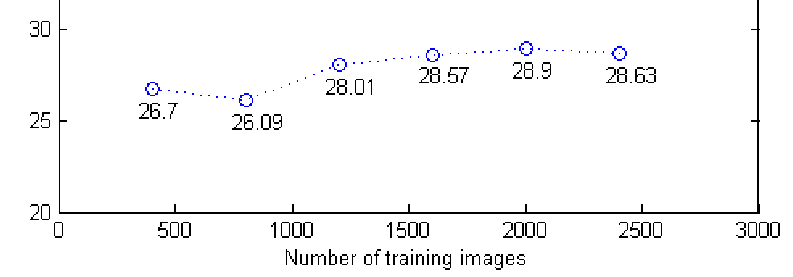}
\rebuttal{\caption{Relationship between mAP and the number of training images for the RR-RR generation setting.}}
\label{fig:number}
\vspace{-1em}
\end{figure}

We first evaluate how variations in low-level cues affect the features generated by the {\small IMGNET} and {\small PASC-FT} networks on the PASCAL VOC2007 dataset. For each experiment, we follow these steps
 (see Figure~\ref{fig:overview}): 1) select cues, 2) generate a batch of synthetic 2D images with those cues, 3) sample positive and negative patches for each class, 4) extract hidden DCNN layer activations from the patches as features, 5) train a classifier for each object category, 6) test the classifiers on real PASCAL images and report mean Average Precision (mAP).
To determine the optimal number of synthetic training images, we computed mAP as a function of the size of the training set, using the RR-RR image generation setting (Table~\ref{tab:color}). Results, shown in Figure~\ref{fig:number}, indicate that the classifier achieves peak performance around 2000 training images, with 100 positive instances for each of the 20 categories, which is the number used for all subsequent experiments.

\vspace{-1em}
\paragraph{Object Color, Texture and Context}
For this experiment, we used 1-2 pose perturbations per view and all views per category. 
We trained a series of detectors on several background and object texture cue configurations, with results shown in Table~\ref{tab:color}. First, as expected, we see that training with synthetic data obtains lower mean AP than training with real data (around 58\% with bounding box regression). Also, the {\small IMGNET} network representation achieves lower performance than the {\small PASC-FT} network, as was the case for real data in~\cite{RCNN}. However, the somewhat unexpected result is that the generation settings \textbf{RR-RR, W-RR, W-UG, RG-RR} with {\small PASC-FT} all achieve comparable performance, despite the fact that \textbf{W-UG} has no texture and no context. Results with real texture but no color in the background (\textbf{RG-RR, W-RR}) are the best. Thus, the {\small PASC-FT} network has learned to be invariant to the color and texture of the object and its background. Also, we note that settings \textbf{RR-UG} and \textbf{RG-UG} achieve much lower performance (6-9 points lower), potentially because the uniform object texture is not well distinguished from the non-white backgrounds. 


For the IMGNET network, the trend is similar, but with the best performing methods being \textbf{RR-RR} and \textbf{RG-RR}. This means that adding realistic context and texture statistics helps the classifier, and thus the IMGNET network is less invariant to these factors, at least for the categories in our dataset. We note that the IMGNET network has seen these categories in training, as they are part of the ILSVRC 1000-way classification task, which explains why it is still fairly insensitive. Combinations of uniform texture with a real background also do not perform well here. Interestingly, \textbf{RG-RR} does very well with both networks, leading to the conclusion that both networks have learned to associate the right context colors with objects.
We also see some variations across categories, e.g., categories like $cat$ and $sheep$ benefit most from adding the object texture cue.


\begin{table*}[t]
\scriptsize  
\begin{center}
\begin{tabular}{p{0.22in} c c c c c c}
& \includegraphics[width=0.7in]{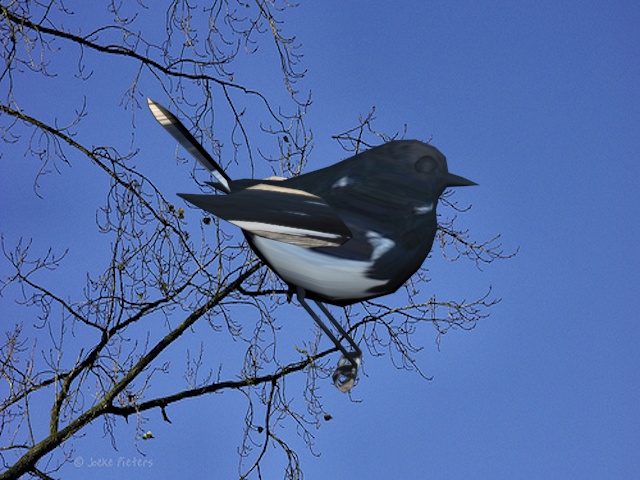}
&\includegraphics[width=0.7in]{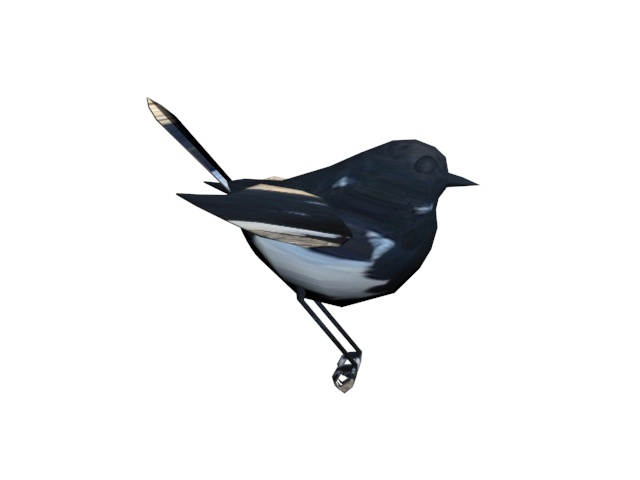}
& \includegraphics[width=0.7in]{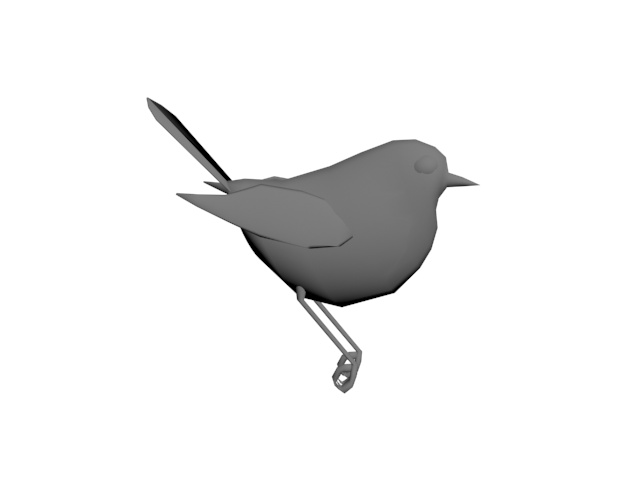}
&\includegraphics[width=0.7in]{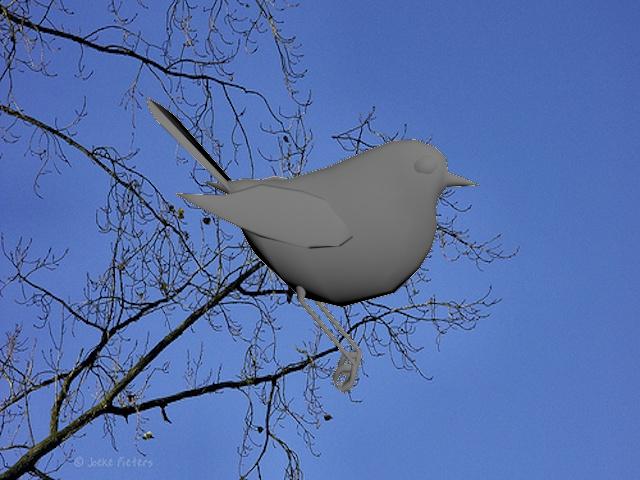}
&\includegraphics[width=0.7in]{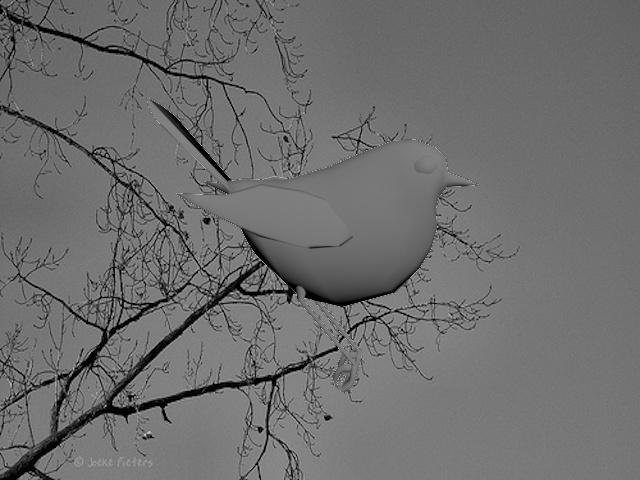}
&\includegraphics[width=0.7in]{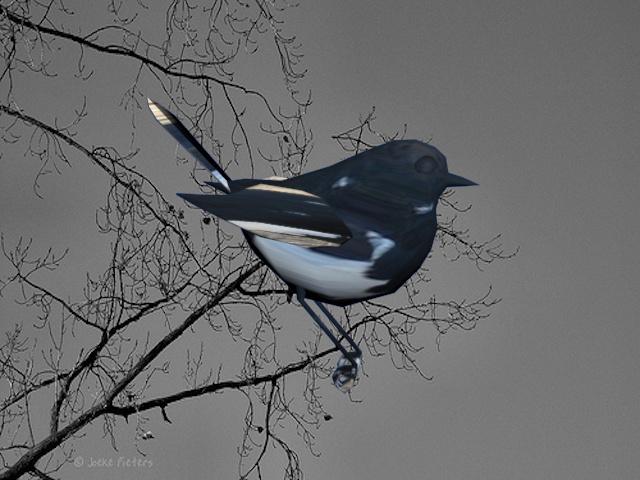}\\
\hline
\end{tabular}
\begin{tabular}{l||>{\centering\arraybackslash}p{0.7in}|>{\centering\arraybackslash}p{0.7in}|>{\centering\arraybackslash}p{0.7in}|>{\centering\arraybackslash}p{0.7in}|>{\centering\arraybackslash}p{0.7in}|>{\centering\arraybackslash}p{0.7in}}
 &\textbf {RR-RR} & \textbf{W-RR}&\textbf{W-UG}  &\textbf{RR-UG}&\textbf{RG-UG}&\textbf{RG-RR}\\ 
\hline
\footnotesize{BG} & Real RGB&White & White &Real RGB&Real Gray&Real Gray\\
\hline
\footnotesize{TX} &Real RGB & Real RGB&Unif. Gray&Unif. Gray&Unif. Gray&Real RGB\\
\hline

\end{tabular}
\begin{tabular}
{ c  ||  p{0.15cm}  p{0.15cm}  p{0.15cm}  p{0.15cm} p{0.15cm} p{0.15cm} p{0.15cm} p{0.15cm}  p{0.15cm} p{0.15cm} p{0.15cm} p{0.15cm} p{0.15cm} p{0.15cm}p{0.15cm} p{0.15cm} p{0.15cm}p{0.15cm} p{0.15cm} p{0.25cm}|p{0.15cm}}
\textsc{PASC-FT}
 &aero& bike&bird&boat&botl	&bus&car&cat&chr	&cow	&tab	&dog	&hse	&mbik	&pers	&plt	&shp	&sofa	&trn	&tv&mAP\\

\hline
\textbf{RR-RR}&50.9 &57.5 &28.3 &20.3 &17.8 &50.1 &37.7 &26.1 &11.5 &27.1 &2.4 &25.3 &40.2 &52.2 &14.3 &11.9 &40.4 &16.3 &15.2 &32.2 &28.9 \\
\textbf{W-RR} &46.5 &55.8 &28.6 &21.7 &21.3 &50.6 &46.6 &28.9 &14.9 &38.1 &0.7 &27.3 &42.5 &53.0 &17.4 &22.8 &30.4 &16.4 &16.7 &43.5 &31.2 \\
\textbf{W-UG}&54.4 &49.6 &31.5 &24.8 &27.0 &42.3 &62.9 &6.6 &21.2 &34.6 &0.3 &18.2 &35.4 &51.3 &33.9 &15.0 &8.3 &33.9 &2.6 &49.0 &30.1\\
\textbf{RR-UG}&55.2 &57.8 &24.8 &17.1 &11.5 &29.9 &39.3 &16.9 &9.9 &35.1 &4.7 &30.1 &37.5 &53.1 &18.1 &9.5 &12.4 &18.2 &2.1 &21.1 &25.2 \\
\textbf{RG-UG}&49.8 &56.9 &20.9 &15.6 &10.8 &25.6 &42.1 &14.7 &4.1 &32.4 &9.3 &20.4 &28.0 &51.2 &14.7 &10.3 &12.6 &14.2 &9.5 &28.0 &23.6 \\
\textbf{RG-RR}&46.5&55.8&28.6&21.7&21.3&50.6&46.6&28.9&14.9&38.1&0.7&27.3&42.5&53.0&17.4&22.8&30.4&16.4&16.7&43.5&31.2\\
\hline
\end{tabular}

\begin{tabular}
{ c  ||  p{0.15cm}  p{0.15cm}  p{0.15cm}  p{0.15cm} p{0.15cm} p{0.15cm} p{0.15cm} p{0.15cm}  p{0.15cm} p{0.15cm} p{0.15cm} p{0.15cm} p{0.15cm} p{0.15cm}p{0.15cm} p{0.15cm} p{0.15cm}p{0.15cm} p{0.15cm} p{0.25cm}|p{0.15cm}}
\textsc{IMGNET}
 &aero& bike&bird&boat&botl	&bus&car&cat&chr	&cow	&tab	&dog	&hse	&mbik	&pers	&plt	&shp	&sofa	&trn	&tv&mAP\\
\hline
\textbf{RR-RR}& 34.3& 34.6& 19.9& 17.1& 10.8& 30.0& 33.0& 18.4& 9.7& 13.7& 1.4& 17.6& 17.7& 34.7& 13.9& 11.8& 15.2& 12.7& 6.3& 26.0& 18.9 \\
\textbf{W-RR}&35.9 &23.3 &16.9 &15.0 &11.8 &24.9 &35.2 &20.9 &11.2 &15.5 &0.1 &15.9 &15.6 &28.7 &13.4 &8.9 &3.7 &10.3 &0.6 &28.8 &16.8 \\
\textbf{W-UG}&38.6 &32.5 &18.7 &14.1 &9.7 &21.2 &36.0 &9.9 &11.3 &13.6 &0.9 &15.7 &15.5 &32.3 &15.9 &9.9 &9.7 &19.9 &0.1 &17.4 &17.1 \\
\textbf{RR-UG}&26.4 &36.3 &9.5 &9.6 &9.4 &5.8 &24.9 &0.4 &1.2 &12.8 &4.7 &14.4 &9.2 &28.8 &11.7 &9.6 &0.7 &4.9 &0.1 &12.2 &11.6 \\
\textbf{RG-UG}&32.7 &34.5 &20.2 &14.6 &9.4 &7.5 &30.1 &12.1 &2.3 &14.6 &9.3 &15.2 &11.2 &30.2 &12.3 &11.4 &2.2 &9.9 &0.5 &13.1 &14.7 \\
\textbf{RG-RR}&26.4 &38.2 &21.0 &15.4 &12.1 &26.7 &34.5 &18.0 &8.8 &16.4 &0.4 &17.0 &20.9 &32.1 &11.0 &14.7 &18.4 &14.8 &6.7 &32.0 &19.3 \\
\end{tabular}
\end{center}
\caption{Detection results on the PASCAL VOC2007 test dataset. Each row is trained on different background and texture configuration of virtual data shown in the top table. In the middle table, the DCNN is trained on ImageNet ILSVRC 1K classification data and finetuned on the PASCAL training data; in the bottom table, the network is not fine-tuned on PASCAL.}
\label{tab:color}
\vspace{-0.1in}
\end{table*}

\begin{figure*}[t]
\centering
\includegraphics[width=0.49\linewidth]{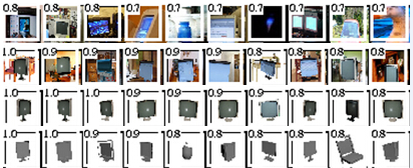}
\includegraphics[width=0.49\linewidth]{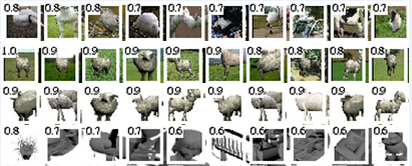}

\caption{\textbf{Top 10 regions with strongest activations for 2 $pool_5$ units} using the method of~\cite{RCNN}. Overlay of the unit's receptive field is drawn in white and normalized activation value is shown in the upper-left corner. For each unit we show results on (top to bottom): real PASCAL images, RR-RR, W-RR, W-UG. See text for further explanation.}
\label{fig:pool5vis}
\vspace{-1em}
\end{figure*}

To explore the lower layers' invariance to color, texture and background, we visualize the patches which have the strongest activations for $pool_5$ units, as shown in Figure \ref{fig:pool5vis}.  The value in the receptive field's upper-left corner is normalized by dividing by max activation value over all units in a channel. The results are very interesting. The unit in the left subfigure fires on patches resembling tv-monitors in real images; when using our synthetic data, the unit still fires on tv-monitors even though the background and texture are removed. The unit on the right fires on white animals on green backgrounds in real and \textbf{RR-RR} images, and continues to fire on synthetic sheep with simulated texture, despite lack of green background. However, it fails on \textbf{W-UG} images, demonstrating its specificity to object color and texture.

\vspace{-1em}
\paragraph{Synthetic Pose}

We also analyse the invariance of CNN features to 3D object pose. Through the successive operations of convolution and max-pooling, CNNs have a built-in invariance to translations and scale. Likewise, visualizations of learned filters at the early layers indicate a built-in invariance to local rotations. Thus while the CNN representation is invariant to slight translation, rotations and deformations, it remains unclear to what extent are CNN representation to large 3D rotations.

For this experiment, we fix the CAD models to three dominant poses: front-view, side-view and intra-view, as shown in Table~\ref{tab:diffpos}. We change the number of views used in each experiment, but keep the total number of synthetic training images (\textbf{RR-RR}) exactly the same, by generating random small perturbations (-15 to 15 degree) around the main view. Results indicate that for both networks adding side view to front view gives a boost, but improvement from adding the third view is marginal. We note that adding some views may even hurt performance (e.g., TV) as the PASCAL test set may not have objects in those views.

\begin{table*}[t]
\scriptsize  
\begin{center}
\includegraphics[width=0.6\linewidth]{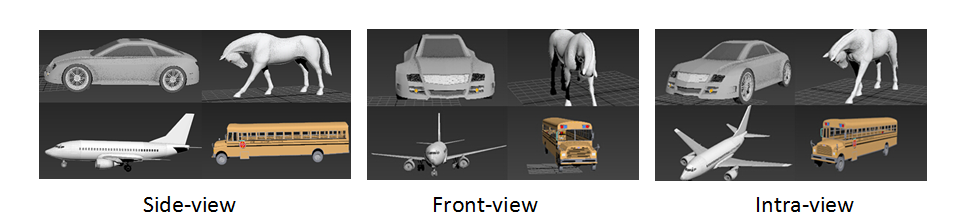} \\
\begin{tabular}
{c  ||  p{0.15cm}  p{0.15cm}  p{0.15cm}  p{0.15cm} p{0.15cm} p{0.15cm} p{0.15cm} p{0.15cm}  p{0.15cm} p{0.15cm} p{0.15cm} p{0.15cm} p{0.15cm} p{0.15cm}p{0.15cm} p{0.15cm} p{0.15cm}p{0.15cm} p{0.15cm} p{0.25cm}|p{0.15cm}}
IMGNET &areo& bike&bird&boat&botl	&bus&car&cat&chr	&cow	&tab	&dog	&hse	&mbik	&pers	&plt	&shp	&sofa	&trn	&tv&mAP\\
\hline
\textbf{front}& 24.9& 38.7& 12.5& 9.3& 9.4& 18.8& 33.6& 13.8& 9.7& 12.5& 2.1& 18.0& 19.6& 27.8& 13.3& 7.5& 10.2& 9.6& 13.8& 28.8& 16.7\\
\textbf{front,side}& 24.3& 36.8& 19.0& 17.7& 11.9& 26.6& 36.0& 10.8& 9.7& 15.5& 0.9& 21.6& 21.1& 32.8& 14.2& 12.0& 14.3& 12.7& 10.1& 32.6& 19.0\\
\textbf{front,side,intra}& 33.1& 40.2& 19.4& 19.6& 12.4& 29.8& 35.3& 16.1& 5.2& 16.5& 0.9& 19.7& 19.0& 34.9& 15.8& 11.8& 19.7& 16.6& 14.3& 29.8& 20.5\\
\hline
\end{tabular}

\begin{tabular}
{c  ||  p{0.15cm}  p{0.15cm}  p{0.15cm}  p{0.15cm} p{0.15cm} p{0.15cm} p{0.15cm} p{0.15cm}  p{0.15cm} p{0.15cm} p{0.15cm} p{0.15cm} p{0.15cm} p{0.15cm}p{0.15cm} p{0.15cm} p{0.15cm}p{0.15cm} p{0.15cm} p{0.25cm}|p{0.15cm}}
{PASC-FT} &aero& bike&bird&boat&botl	&bus&car&cat&chr	&cow	&tab	&dog	&hse	&mbik	&pers	&plt	&shp	&sofa	&trn	&tv&mAP\\
\hline
\textbf{front}& 41.8& 53.7& 14.5& 19.1& 11.6& 42.5& 40.4& 25.5& 9.9& 24.5& 0.2& 29.4& 37.4& 47.1& 14.0& 11.9& 18.9& 12.7& 22.6& 38.8& 25.8\\
\textbf{front,side}& 45.6& 50.2& 24.4& 28.8& 17.4& 51.9& 41.8& 24.5& 7.2& 27.9& 9.2& 23.1& 37.0& 51.3& 17.8& 13.2& 28.6& 18.9& 9.3& 37.8& 28.3\\
\textbf{front,side,intra}& 54.2& 55.5& 22.7& 27.0& 20.5& 52.6& 40.1& 26.8& 8.1& 27.3& 2.3& 30.6& 36.6& 53.3& 17.8& 14.2& 34.1& 26.4& 19.3& 37.5& 30.3\\
\end{tabular}
\end{center}
\caption{Results of training on different synthetic views. The CNN used in the top table is trained on ImageNet-1K classification, the CNN in the bottom table is also finetuned on PASCAL 2007 detection.}
\label{tab:diffpos}
\vspace{-0.1in}
\end{table*}

\vspace{-1em}
\paragraph{Real Image Pose}

We also test view invariance on real images. We are interested here in objects whose frontal view presentation differs significantly (ex: the side-view of a horse vs a frontal view). To this end, we selected 12 categories from the PASCAL VOC training set which match this criteria. Held out categories included rotationally invariant objects such as bottles or tables. Next, we split the training data for these 12 categories to prominent side-view and front-view, as shown in Table~\ref{tab:realpose}.


We train classifiers exclusively by removing one view (say front-view) and test the resulting detector on the PASCAL VOC test set containing both side and front-views.We also compare with random view sampling. Results, shown in Table~\ref{tab:realpose}, point to important and surprising conclusions regarding the representational power of the CNN features. Note that mAP drops by less than $2\%$ when detectors exclusively trained  by removing either view are tested on the PASCAL VOC test set. Not only are those detectors never presented with the second view, but they are also trained with approximately half the data. While this invariance to large and complex pose changes may be explained by the fact the CNN model was itself trained with both views of the object present, and subsequently fine-tuned with both views again present, the level of invariance is nevertheless remarkable. In the last experiment, we reduce the fine-tuning training set by removing front-view objects, 
and note a larger mAP drop of $5$ points (8\%), but much less than one may expect. We conclude that, for both networks, the representation groups together multiple views of an object. 

\begin{table*}[t]
\scriptsize
\begin{center}
\includegraphics[width=0.6\linewidth]{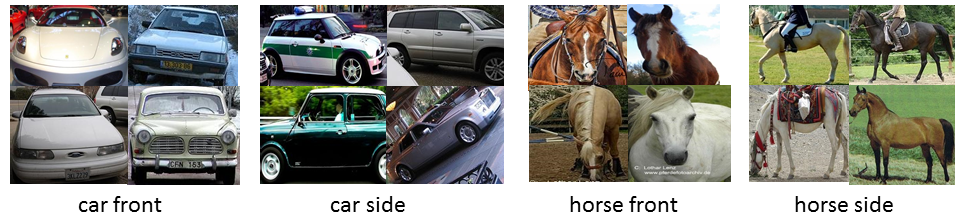} \\
\begin{tabular}
{c |c ||  p{0.25cm}  p{0.25cm}  p{0.25cm}  p{0.25cm} p{0.25cm} p{0.25cm} p{0.25cm} p{0.25cm}  p{0.25cm} p{0.25cm} p{0.25cm} p{0.40cm} |p{0.35cm}}

Net & Views &aero& bike&bird&bus&car&cow	&dog	&hrs	&mbik	&shp	&trn	&tv&mAP\\
\hline
PASC-FT & \textbf{all}&64.2&69.7&	50&	62.6&	71&	58.5&	56.1&	60.6&	66.8&	52.8&	57.9&	64.7&	61.2\\
PASC-FT &\textbf{-random}& 62.1&70.3&49.7&61.1&70.2&54.7&55.4&61.7&67.4&55.7&57.9&64.2&60.9\\
PASC-FT &\textbf{-front}&61.7&67.3&45.1&58.6&70.9&56.1&55.1&	59.0&66.1&54.2&53.3&61.6&59.1\\
PASC-FT & \textbf{-side}&62.0&70.2&48.9&61.2&70.8&57.0&53.6&59.9&65.7&53.7&58.1&64.2&60.4\\

PASC-FT(-front) & \textbf{-front}	&59.7&63.1&42.7&55.3&64.9&54.4&54.0&56.1&64.2&55.1&47.4&60.1&56.4\\
\hline
\end{tabular}


\end{center}
\caption{Results of training on different real image views. '-' represent removing a certain view. Note that the mAP is only for a subset of Pascal Dataset.
}
\label{tab:realpose}
\vspace{-0.1in}
\end{table*}

\vspace{-1em}
\paragraph{3D Shape}


Finally, we experiment with reducing intra-class shape variation by using fewer CAD models per category. We otherwise use the same settings as in the \textbf{RR-RR} condition with PASC-FT. 
 From our experiments, we find that the mAP decreases by about 5.5 points from 28.9\% to 23.53\% when using only a half of the 3D models. This shows a significant boost from adding more shape variation to the training data, indicating less invariance to this factor.

\subsection{Few-Shot Learning Results on PASCAL}

To summarize the conclusions from the previous section, we found that DCNNs learn a significant amount of invariance to texture, color and pose, and less invariance to 3D shape, if trained (or fine-tuned) on the same task. If not trained on the task, the degree of invariance is lower. Therefore, when learning a detection model for a new category with no or limited labeled real data available, it is advantageous to simulate these factors in the synthetic data.

In this section, we experiment with adapting the deep representation to the synthetic data. We use all available 3D models and views, and compare the two generation settings that produced the best results (\textbf{RR-RR, RG-RR} in Table~\ref{tab:color}). Both of these settings use realistic backgrounds, which may have some advantages for detection.
In particular, visualizations of the positive training data show that a white background around the objects makes it harder to sample negative training data via selective search, as most of the interesting regions are on the object.

As before, we simulate the zero-shot learning situation where the number of labeled real images for a novel category is zero, however, here we also experiment with having a small number of labeled real images. For every category, we randomly select 20 (10, 5) positive training images to form datasets $R_{20}$ ($R_{10}$, $R_{5}$). The sizes of final datasets are 276 (120, 73); note that there are some images which contain two or more positive bounding boxes. The size of the virtual dataset (noted as $V_{2k}$) is always 2000 images. We pre-train on Imagenet ILSVRC (IMAGENET network) and fine-tune on $V_{2k}$ to get the {\small VCNN} network, then train SVM classifiers on both $R_{x}$+$V_{2k}$.

\noindent\textbf{Baselines.} We use datasets $R_{x}$ ($x=20,10,5$) to train the RCNN model, and $R_{x}$+$V_{2k}$ to train the Fast Adaptation method described in \cite{BMVC}. 
The RCNN is pre-trained on Imagenet ILSVRC, however it is not fine-tuned on detection on $R_{5}$ and $R_{10}$ as data is very limited. 

\begin{figure}[tb]
\centering
\includegraphics[width=0.65\linewidth]{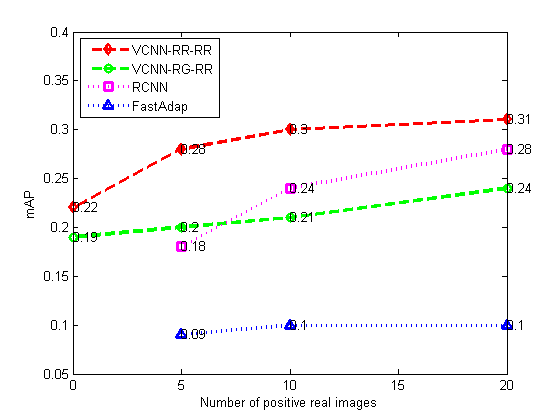}
\caption{Detection results of the proposed VCNN on PASCAL. When the real annotated images are limited or not available, eg. for a novel category, VCNN performs much better than RCNN and the Fast Adaptation method.}
\label{results}
\end{figure}

\begin{table*}[t]
\scriptsize
\begin{center}

\begin{tabular}
{c  || p{0.03cm} p{0.03cm} p{0.03cm} p{0.03cm} p{0.03cm} p{0.03cm} p{0.03cm} p{0.03cm}  p{0.03cm} p{0.03cm} p{0.03cm} p{0.03cm} p{0.03cm} p{0.03cm} p{0.03cm} p{0.03cm} p{0.03cm} p{0.03cm} p{0.03cm} p{0.03cm} p{0.03cm} p{0.03cm} p{0.03cm} p{0.03cm} p{0.03cm} p{0.03cm} p{0.03cm} p{0.03cm} p{0.03cm} p{0.03cm} p{0.16cm}   |p{0.15cm}}

Training&bp&bk&bh&bc&bt&ca&dc&dl&dp&fc&hp&kb&lc&lt&mp&mt&ms&mg&pn&pe&ph&pr&pj&pn&rb&rl&sc&sp&st&td&tc&mAP\\
\hline
WEBCAM&81& 91& 65& 35& 9& 52& \textbf{84}& 30& 2& \textbf{33}& 67& 37& 71& 14& 21& \textbf{54}& 71& 38& 26& \textbf{19}& 41& 58& 64& \textbf{16}& 10& 11& 32& 1& 18& 29& 26& 39\\
V-GRAY& 81& 93& \textbf{65}& \textbf{35}& \textbf{30}& 17& 84& 30& 2& 33& \textbf{67}& 37& 71& 14& 21& 17& 24& 9& 26& 9& 4& 58& 54& 16& 10& 11& \textbf{32}& 1& 18& \textbf{29}& 26&33\\
V-TX&\textbf{89}& \textbf{94}& 40& 32& 20& \textbf{81}&83& \textbf{48}& \textbf{15}& 19& 72& \textbf{66}& \textbf{78}& \textbf{18}& \textbf{77}& 49& \textbf{75}& \textbf{73}& \textbf{26}& 17& \textbf{41}& \textbf{64}& \textbf{77}& 15& \textbf{10}& \textbf{15}& 29& \textbf{29}& \textbf{29}& 24& \textbf{31}& \textbf{46}\\

\hline
\end{tabular}

\end{center}
\caption{Detection results of the proposed VCNN on the 31 object categories in the Office dataset. The test data in these experiments are (real) images from the Amazon domain. We compare to training on the real training images from the Webcam domain (top row). Our model was trained on V-GRAY and V-TX, representing virtual images with uniform gray texture and  real texture, respectively. The results clearly demonstrate that when the real training data is mismatched from the target domain, synthetic training can provide a significant performance boost for real-image detection.  
}
\label{tab:officeresult}
\vspace{-0.1in}
\end{table*}

\begin{figure}[t]
\centering
\includegraphics[width=0.9\linewidth]{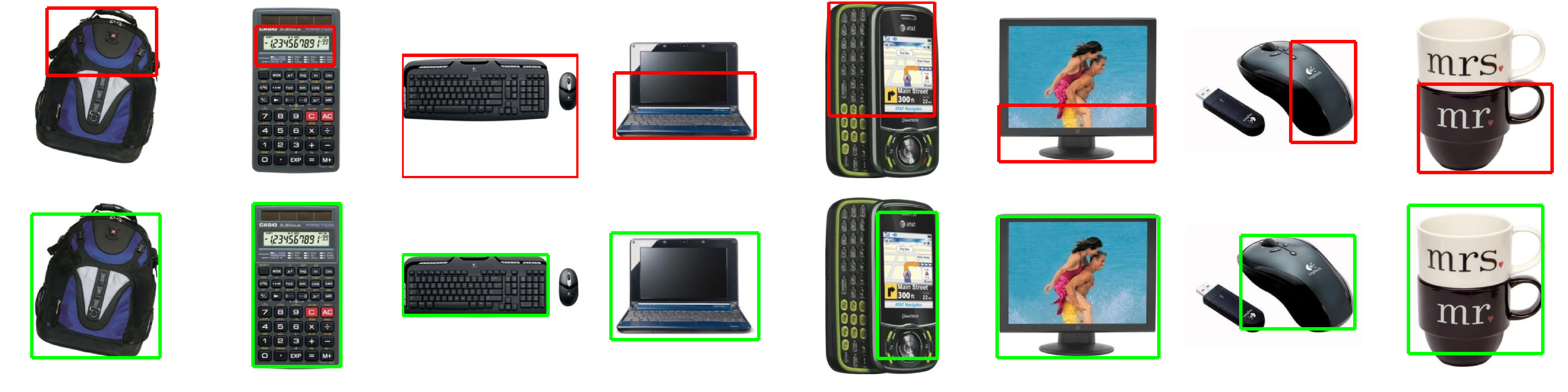}
\caption{Detections on the Amazon domain in Office, showing examples where our synthetic model (second row, green bounding box) improves localization compared to the model trained on real Webcam images (first row, red bounding box).}
\label{fig:detection_ex}
\end{figure}


\noindent\textbf{Results.} The results in Figure~\ref{results} show that when the number of real training images is limited, our method (VCNN) performs better than traditional RCNN. The VCNN also significantly outperfoms the Fast-Adapt method, which is based on HOG features. We also confirm that our proposed \textbf{RR-RR} data synthesis methodology is better than not simulating background or texture. In partcular, fine-tuning on virtual \textbf{RR-RR} data boosts mAP from 18.9\% (Table~\ref{tab:color}) to 22\% without using any real training examples, and to 28\% with 5 real images per category, a 10\% absolute improvement over RCNN. We also notice that the results for \textbf{RG-RR} are much lower than \textbf{RR-RR}, unlike the results in the fixed-feature experiment. This may be explained by the fact that \textbf{RG-RR} with selective search generates many sub-regions without color, and using these regions to do fine-tuning probably decreases the CNN's ability to recognize realistic color objects. 


Note that the VCNN trained with 10 real images per category (200 total) is also using the approximately 900 real images of texture and background. However, this is still much fewer than the 15588 annotated bounding boxes in the PASCAL training set, and much easier to collect as only the texture images (about 130) need bounding box annotation. Yet the obtained 31\% mAP is comparable to the 33\% mAP achieved by the DPM (without context rescoring) trained on the full dataset. This speaks to the power of transferring deep representations and suggests that synthetic CAD data is a promising way to avoid tedious annotation for novel categories.
We emphasize that there is a significant boost due to adapting the features on synthetic data via fine-tuning, showing that adapted features are better than fixed features, but only for the \textbf{RR-RR} generation settings. 

\subsection{Results on Novel Domains}

When the test images come from a different visual domain (or dataset) than the training images, we expect the performance of the detector to degrade due to dataset bias~\cite{office}. In this experiment, we evaluate the benefit of using synthetic CAD data to improve performance on novel real-image domains.
We use part of the Office dataset ~\cite{office}, which has the same 31 categories of common objects (cups, keyboards, etc.) in each domain, with Amazon images as the target testing domain (downloaded from amazon.com) and Webcam images as the training domain (collected an office environment).

To generate synthetic data for the categories in the Office Dataset, we downloaded roughly five 3D models for each category. The data generation method is the same as the experiments for PASCAL, expcept that we use the original texture on the 3D models for this experiment, considering that the texture of the objects in Office dataset is simpler. 
We compare two generation settings, \textbf{V-GRAY} and \textbf{V-TX}, representing virtual images with uniform gray texture and  real texture, respectively. The background for both settings is white, to match the majority of Amazon domain backgrounds.
We generate 5 images for each model, producing 775 images in total.  We use the synthetic images to train our VCNN deep detector and test it on the Amazon domain (2817 images).

\noindent \textbf{Baseline}
We train a baseline real-image deep detector on the Webcam domain (total of 795 images) and also test it on images in the Amazon domain.

\noindent \textbf{Results}
The results are shown in Table~\ref{tab:officeresult}. The mean AP for VCNN with \textbf{V-TX} is 46.25\% versus 38.91\% for the deep detector trained on the Webcam domain, a significant boost in performance. The \textbf{V-GRAY} setting does considerably worse. This shows the potential of synthetic CAD training in dataset bias scenarios.

In Figure~\ref{fig:detection_ex}, we show some examples where the object is not detected by the detector trained on Webcam, but detected perfectly by the our VCNN model. To obtain these results we selected the bounding box with the highest score from about 2000 region proposals in each image.
%

%% file: conclusion.tex
\section{Conclusion}
This paper demonstrated that synthetic CAD training of modern deep CNNs object detectors can be successful when real-image training data for novel objects or domains is limited.
We investigated the sensitivity of convnets to various low-level cues in the training data: 3D pose, foreground texture and color, background image and color. To simulate these factors we used synthetic data generated from 3D CAD models.
Our results demonstrated that the popular deep convnet of \cite{alexnet}, fine-tuned for the detection task, is indeed largely invariant to these cues. Training on synthetic images with simulated cues lead to similar performance as training on synthetic images without these cues. However, if the network is not fine-tuned for the task, its invariance is diminished. Thus, for novel categories, adding synthetic variance along these dimensions and fine-tuning the layers proved useful. 

Based on these findings, we proposed a new method for learning object detectors for new categories that avoids the need for costly large-scale image annotation. 
This can be advantageous when one needs to learn a detector for a novel object category or instance, beyond those available in labeled datasets. We also showed that our method outperforms detectors trained on real images when the real training data comes from a different domain, for one such case of domain shift. These findings are preliminary, and further experiments with other domains are necessary. 

\section{Acknowledgements}
We thank Trevor Darrell, Judy Hoffman and anonymous reviewers for their suggestions. This work was supported by the NSF Award No. 1451244.